\documentclass[conference]{IEEEtran}
\usepackage{makecell}
\IEEEoverridecommandlockouts
\usepackage{cite}
\usepackage{amsmath,amssymb,amsfonts}
\usepackage{algorithmic}
\usepackage{graphicx}
\usepackage{textcomp}
\usepackage{xcolor}
\def\BibTeX{{\rm B\kern-.05em{\sc i\kern-.025em b}\kern-.08em
    T\kern-.1667em\lower.7ex\hbox{E}\kern-.125emX}}
\begin{document}

\title{Outdoor Monocular Depth Estimation: A Research Review}

\author{\IEEEauthorblockN{Pulkit Vyas, Chirag Saxena, Anwesh Badapanda, Anurag Goswami}
\IEEEauthorblockA{\textit{School of Computer Science and Technology} \\
\textit{Bennett University}\\
Greater Noida, India \\
\{e19cse284, e19cse124, e19cse058, anurag.goswami\} @bennett.edu.in}
}

\maketitle

\begin{abstract}
Depth estimation is an important task, applied in various methods and applications of computer vision. While the traditional methods of estimating depth are based on depth cues and require specific equipment such as stereo cameras and configuring input according to the approach being used, the focus at the current time is on a single source, or monocular, depth estimation. The recent developments in Convolution Neural Networks along with the integration of classical methods in these deep learning approaches have led to a lot of advancements in the depth estimation problem. The problem of outdoor depth estimation, or depth estimation in wild, is a very scarcely researched field of study. In this paper, we give an overview of the available datasets, depth estimation methods, research work, trends, challenges, and opportunities that exist for open research. To our knowledge, no openly available survey work provides a comprehensive collection of outdoor depth estimation techniques and research scope, making our work an essential contribution for people looking to enter this field of study.
\end{abstract}


\vspace{12pt}

\section{Introduction}
The ability to detect, classify, segment, and reconstruct outdoor and long-range distance scenes is an important requirement for computer vision techniques in application domains and use-cases of autonomous vehicles, robotic systems, 3D architectural modeling, terrestrial surveys, and AR/VR \cite{b1}. Extracting depth information from outdoor scenes becomes a task of utmost importance since that provides a lot of context about the spatial and logical relationship between the different entities present in them. Techniques such as robust point-cloud-based methods or stereo-based methods might seem to be a viable solution to this problem, and in fact, there has been a lot of research being done on them \cite{b2}\cite{b3}. The limiting factors of mass adoption of these technologies are the requirement of specific equipment and the restriction of usage of the input data from them. Most applications cannot account for the need for sensors such as LiDAR needed by these methods, moreover, the applications might include computer vision techniques such as object detection, tracking, and segmentation that would require 2D images instead of sensor data. Thus the requirement for monocular depth estimation systems that work for outdoor data is quite evident. 

\begin{figure}[htbp]
\centerline{\includegraphics[scale=0.08]{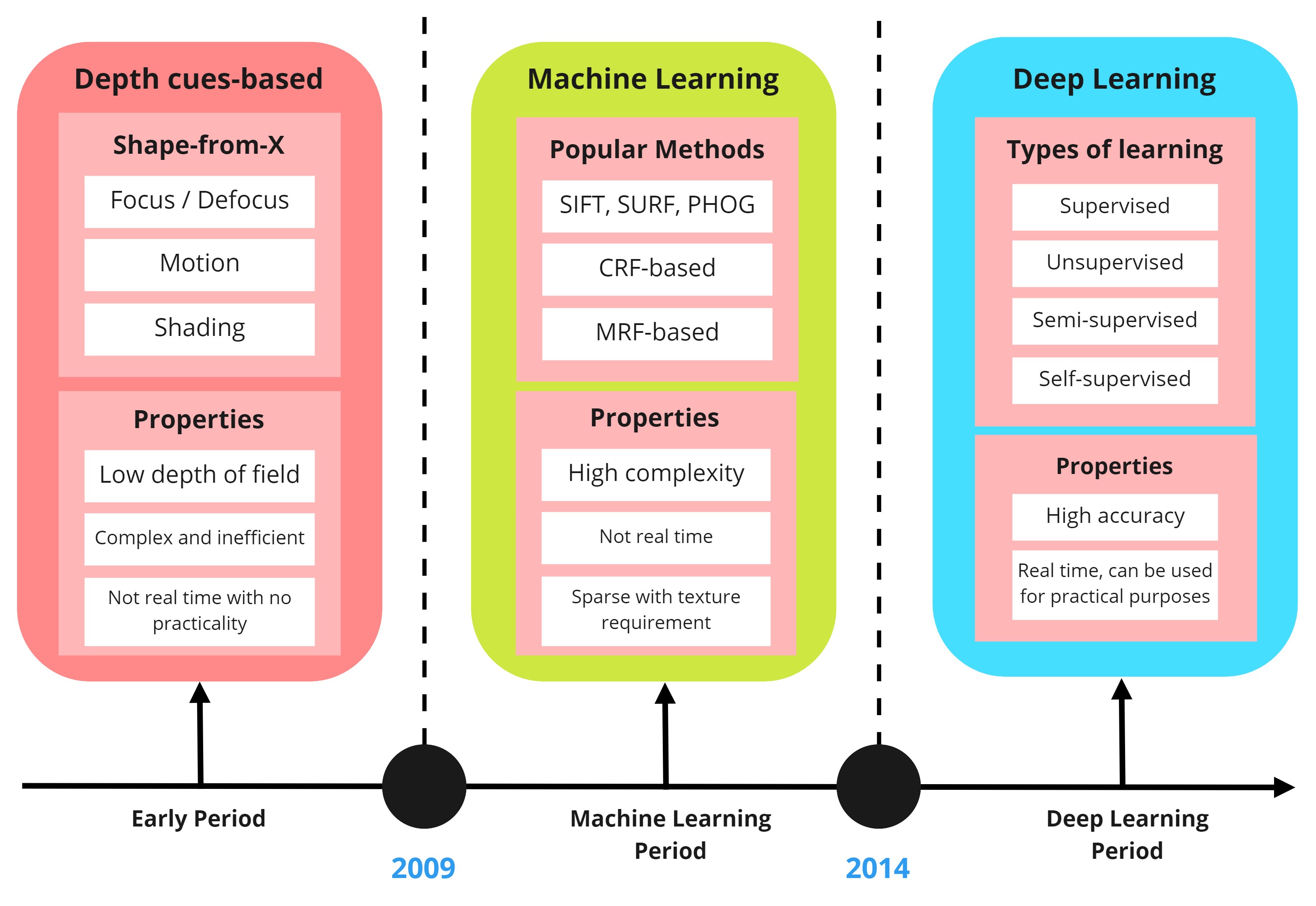}}
\caption{Evolution of techniques for depth estimation over the years. Inspired from \cite{b7}}
\label{fig:image1}
\end{figure}

The biggest challenge with outdoor monocular depth estimation is the lack of perspective changes in the scenes and frames of input image data \cite{b17}\cite{b22}. This happens because when the subject of the scene is much bigger than the focal view size of the camera then the overall change between the elements captured in a set of frames would be much less compared to the same camera being used in an indoor setting. Classical techniques based on depth cues from motion and texture thus fail because of the lack of dynamic nature \cite{b4}\cite{b5}\cite{b6}. Fig. \ref{fig:image1} gives an overview of the evolution of techniques used for depth estimation.

This issue is accompanied by the lack of publicly available outdoor datasets for the training and evaluation of models \cite{b17}. Depth estimation datasets are majorly used in research work in the fields of robotics and 3D reconstruction, both of which are predominantly indoor-based activities in their current form and industrial usage. Due to this, the number of outdoor datasets is very less, with the ones targeting very long-range data being very few. 

The current research in the direction of outdoor depth estimation is driven by the need of capturing or synthetically generating outdoor scene and their depth data, building deep learning-based models and integrating traditional methods wherever applicable, and, applying various learning approaches and improving upon them \cite{b7}. 

This paper concentrates on the research of monocular depth estimation in the context of outdoor scenes, surveying the developments and trends in deep learning-based approaches over the last few years. We also provide a look at the limitations of current research work and highlight future research directions. The rest of the paper is structured in the following way: Section 2 summarizes the datasets that are either built solely with outdoor scenes or have some components of them, section 3 introduces the previous and contemporary deep learning models proposed over the years, section 4 discusses the variety of techniques used for training deep learning models for the task of depth estimation, section 5 outlines the challenges and trends of outdoor depth estimation, we thus conclude in section 6.

\vspace{12pt}

\section{Methodology}
A literature review is supposed to integrate existing information, detect issues related to bias or problematic trends, and identify gaps in the literature of a particular field. Since our study aims to survey and summarize the previous as well as contemporary work in the domain of outdoor depth estimation, the methodology that we follow was to do a backward snowballing of research work basis on the recent works accepted at popular conferences in computer vision like CVPR and ICCV, rather than discovering the papers with the arbitrary search of keywords. Instead, we first collected a suite of relevant keywords and used them to conduct the survey search using Google Scholar, Scopus, and Connected Paper's search engines. This allowed us to find the patterns of research, development, and industry adoption over the years and the kind of approaches that were involved in them. 

The “ill-posed” nature of the problem of monocular depth estimation is especially highlighted when done in outdoor settings. This creates a significant difference in the kind of methods used by indoor depth estimation solutions and the outdoor ones. Due to this, we divide the datasets, deep learning approaches as well as training paradigms in such a way to point out the progress and limitations of current works in this context. 

The inclusion and exclusion of papers in this work are based on the criteria defined in Table 1. 

\begin{table}[htbp]
\caption{Inclusion and Exclusion Criterion}
  \begin{center}
    \begin{tabular}{| l | c |}
      \hline
      \thead{Criterion} & \thead{Included} \\
      \hline
      \makecell{Papers that are focused on outdoor depth  \\ estimation} & Yes  \\
      \hline
      \makecell{Papers using monocular sources, stereo \\ sources, synthetic data, or panoramic images \\ as datasets} & Yes  \\
      \hline
      \makecell{Papers that exclusively use indoor depth \\ estimation datasets} & No  \\
      \hline
      \makecell{Papers that only use private datasets} & No  \\
      \hline
      \makecell{Papers that use depth estimation as the \\ input rather than assisting or improving \\ the results of an MDE system} & No  \\
      \hline
    \end{tabular}
    \label{tab1}
  \end{center}
\end{table}

\vspace{12pt}

\section{Datasets}
The work of \cite{b8} gives a comprehensive review of the publicly available datasets for monocular depth estimation. Still, the majority of datasets meant for monocular depth estimation are created in indoor settings \cite{b9}. This is primarily because it is much easier, both in terms of equipment cost and human effort, to create a dataset indoors as compared to taking the various steps of outdoor data collection, annotation, and pre-processing in consideration and creating a novel dataset based on that. Yet, there are a few notable datasets available for the outdoor depth estimation, here we discuss these datasets into the following categories: generic outdoor datasets, panoramic datasets, and, generative methods. There are a few other datasets like \cite{b61} that are adapted from other domains such as semantic segmentation for the task of monocular depth estimation but are not inherently made for these tasks and so they don't contain depth maps as GT. Depth is extracted in these using artificial methods making the data not ideal for training MDE models but still, a lot of research utilizes these. We do not discuss such datasets in this section but reference them when discussing methods that use them in subsequent sections.

\vspace{7pt}

\subsection{Generic outdoor datasets}
The collection of data is generally done either using moving vehicles or pictures of buildings and scenes with a limited focal view at a short distance (less than 100 meters). This category of depth estimation datasets include the KITTI dataset \cite{b9}, Make3D dataset \cite{b10}, Newer College dataset \cite{b11}, Megadepth dataset \cite{b12}, DIODE \cite{b13} and DrivingStereo dataset \cite{b14}. The subsequent subsections give a brief on the two of the most used of these datasets:

\vspace{9pt}

\subsubsection{KITTI}
\begin{figure}[htbp]
\centerline{\includegraphics[scale=0.35]{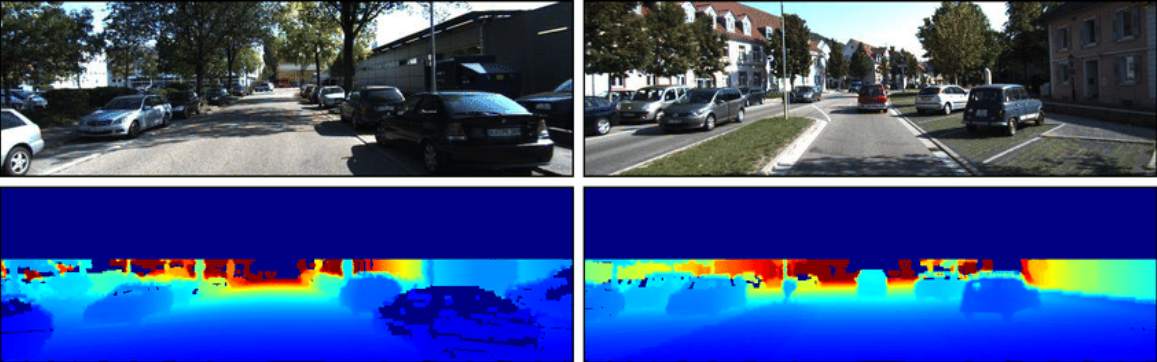}}
\caption{Sections of the KITTI dataset where scenes are captured in an outdoor setting}
\label{fig:image2}
\end{figure}

The KITTI dataset \cite{b9} is the most used reference with exterior scenes taken from a moving vehicle. There are two main divisions used to estimate monocular depth in outdoor environments. One is a training/test set with 23,488 pairs of training images and 697 test images. The other is an official crash with 42,949 pairs of training images, 1000 validation images, and 500 test images. For the official division, the true depth map for the test images is retained with the benchmark reviewer to test models against new data. Fig. \ref{fig:image2} provides a few examples of the outdoor parts of the KITTI dataset.

\vspace{9pt}

\subsubsection{Make3D}
\begin{figure}[htbp]
\centerline{\includegraphics[scale=0.7]{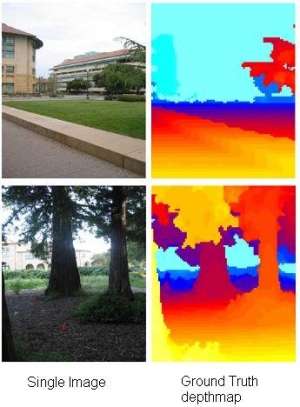}}
\caption{The Make3D Laser + Image dataset}
\label{fig:image3}
\end{figure}

The Make3D dataset \cite{b10} is another outdoor dataset for deep learning-based monocular depth estimation. The Make3D dataset includes cityscapes and natural landscapes captured during the daytime, with the collection of depth maps done by laser scanners. The dataset contains a total of 53 pairs of RGBD images, of which 400 pairs are used for training and 13 pairs are used for testing. The native RGB image resolution is 2272 x 170 and the depth map resolution is 55 x 305 pixels. Fig. \ref{fig:image3} gives an overlook of the Make3D dataset.

\vspace{9pt}

\subsection{Panoramic datasets}
\begin{figure}[htbp]
\centerline{\includegraphics[scale=0.30]{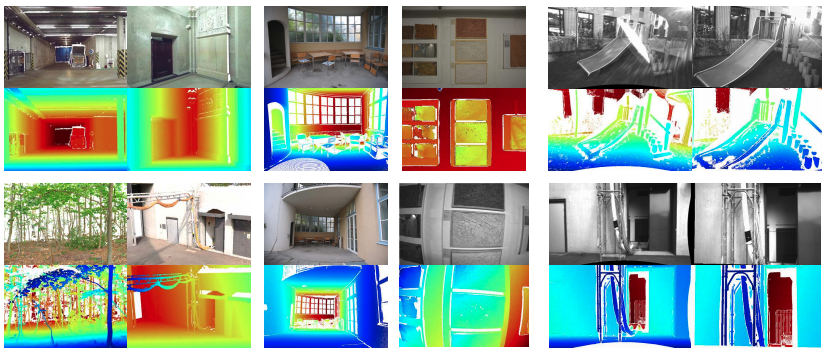}}
\caption{The ETH3D dataset with panoramic images and corresponding depth maps}
\label{fig:image4}
\end{figure}

These datasets are collected using panoramic cameras and meant for usage by large-focal length input devices as well as 3D cameras. Panoramic datasets have been used in a lot of recent research work \cite{b15}\cite{b16}. Still, the number of outdoor panoramic datasets is quite low \cite{b17}\cite{b18}, resulting in a lack of research on the depth estimation for the same. The noteworthy panoramic datasets are Multi-FoV (Urban Canyon) dataset \cite{b19}, ETH3D \cite{b20}, and Forest Virtual \cite{b21}. A few snippets from the dataset are shown in Fig. \ref{fig:image4}.

\vspace{9pt}

\subsection{Generative methods}
Depending upon the domain of the application that particular research work is focusing on, the availability of the specific datasets might be a roadblock. The works of \cite{b22} and \cite{b23} adapt existing datasets such as the KITTI dataset or public image and 3D data from sources like Google Maps to generate novel datasets that have very long-range scenes and 360 panoramic scenes, respectively.

\vspace{9pt}

\subsubsection{FarSight}
\begin{figure}[htbp]
\centerline{\includegraphics[scale=0.35]{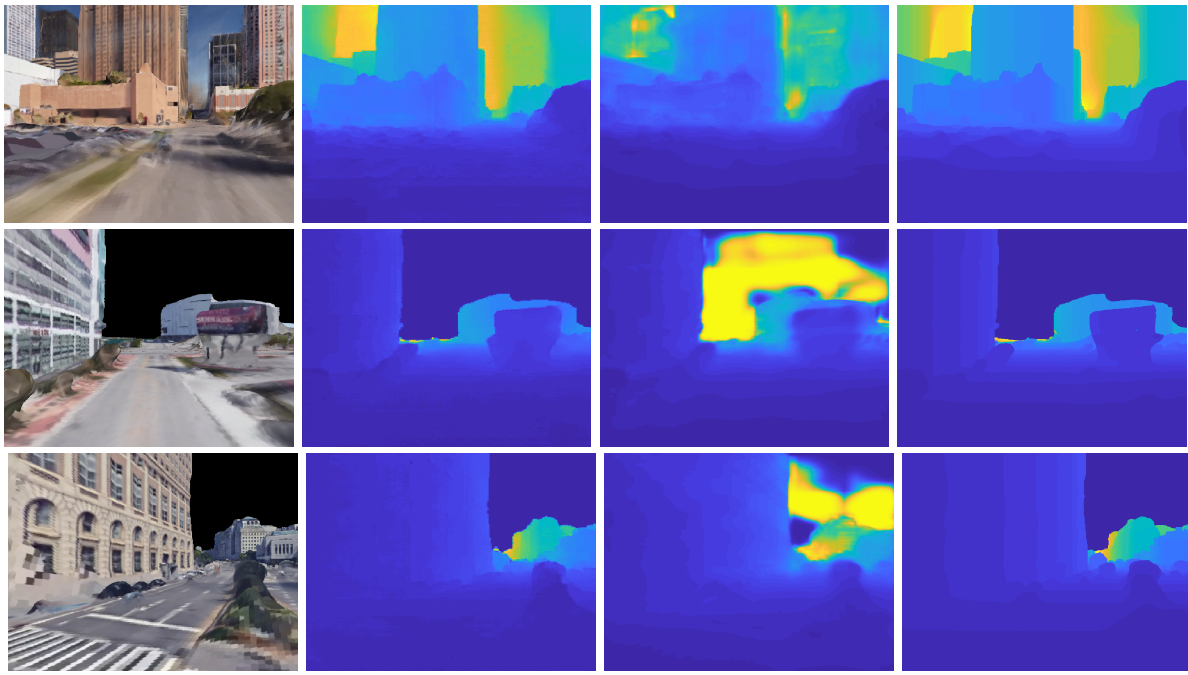}}
\caption{From left to right: the 3D models of the cityscapes extracted from Google Earth, depth prediction using the paper's GAN model, depth prediction using the paper's CNN model, ground truth depth maps}
\label{fig:image5}
\end{figure}

This work \cite{b22} is a strategy for generating very-long range outdoor images along with annotated depth maps. They use a new strategy for generating aggregate data of long-range ground truth depths. They used images from Google Earth to recreate large-scale 3D models of different cities at the appropriate scale. The acquired archive of 3D models and associated RGB views and their long-range depth rendering is used as training data for depth prediction. Fig. \ref{fig:image5} shows the resulting images and depth maps from following this process.

\vspace{9pt}

\subsubsection{KITTI to panoramic dataset adaption}
\begin{figure}[htbp]
\centerline{\includegraphics[scale=0.4]{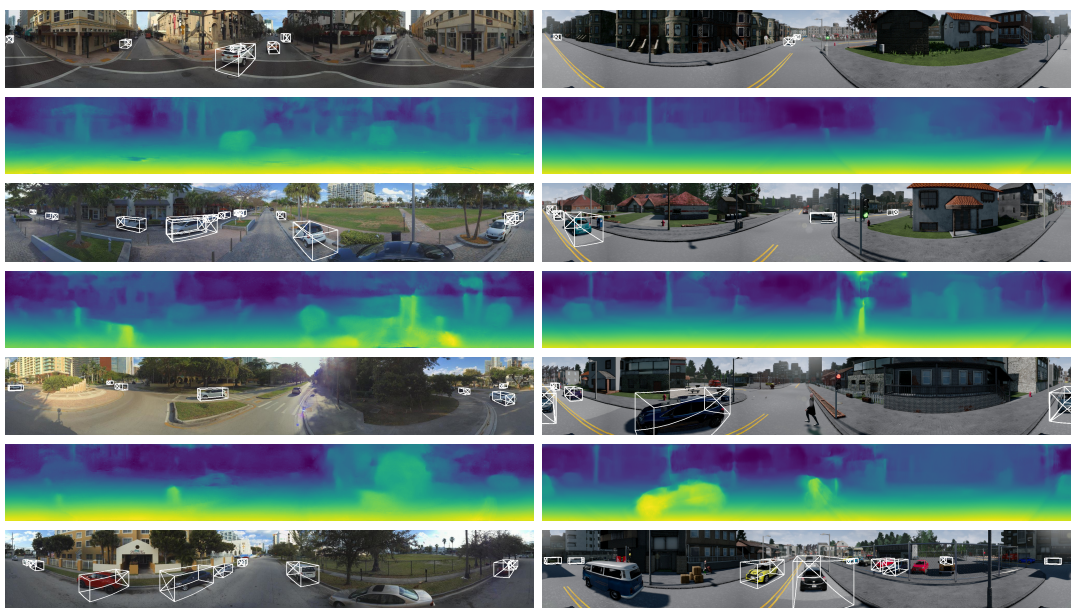}}
\caption{Depth recovery from panoramic imagery using the approach described in \cite{b23}}
\label{fig:image6}
\end{figure}

In this approach \cite{b23}, a two-step process is used where the images from the source, the KITTI dataset, are transformed into a replicated version of their target domain which requires images to be captured by 360-degree FOV cameras. The first step is the style transfer of 360-degree images to the images in the target dataset via a learning-based approach \cite{b24}. The second step involves reprojection of the resulting images to the required format along with the relevant annotations. The generated depth maps can be visualized in Fig. \ref{fig:image6}

\vspace{12pt}

\section{Deep Learning Approaches}

\subsection{Convolutional Neural Networks}
\begin{figure}[htbp]
\centerline{\includegraphics[scale=0.30]{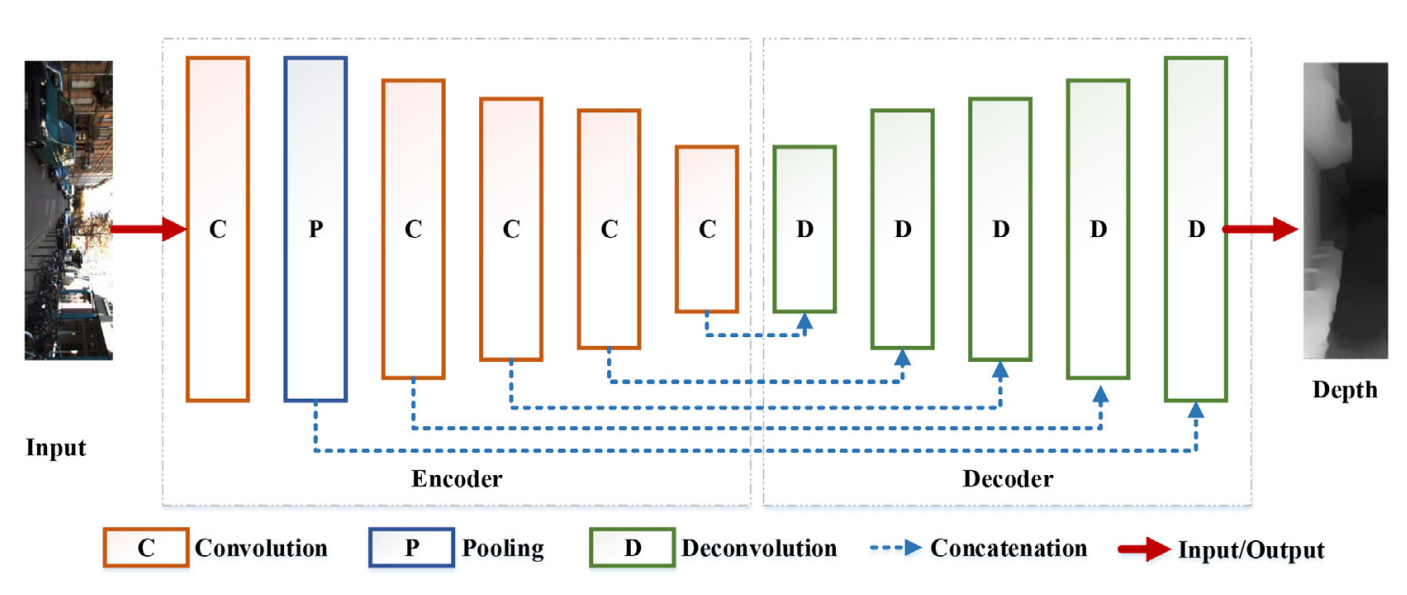}}
\caption{The general pipeline of deep learning for monocular depth estimation using CNNs. Source: \cite{b7}}
\label{fig:image7}
\end{figure}

Convolutional neural networks are primarily used on images with the major components - convolutional layers, pooling layers (max pooling and average pooling), and activation functions - which together allow these networks in learning 2D spatial features of input images. In the context of depth estimation, CNNs are used for extracting depth features from images, reducing the size of these extractions using the pooling layers, and reconstructing the depth maps using their activation functions and FC layers. CNN-based approaches for outdoor depth estimation, mostly for KIITI dataset experiments, include Convolutional Neural Fields \cite{b25} AdaBins \cite{b26}, FarSight \cite{b22}, and, Unsupervised CNN \cite{b27}, all of which use CNNs as the backend for the main depth estimation network. Fig. \ref{fig:image7} shows the steps involved in predicting depth maps using convolutional layers.

\vspace{9pt}

\subsection{Recurrent Neural Networks}
\begin{figure}[htbp]
\centerline{\includegraphics[scale=0.35]{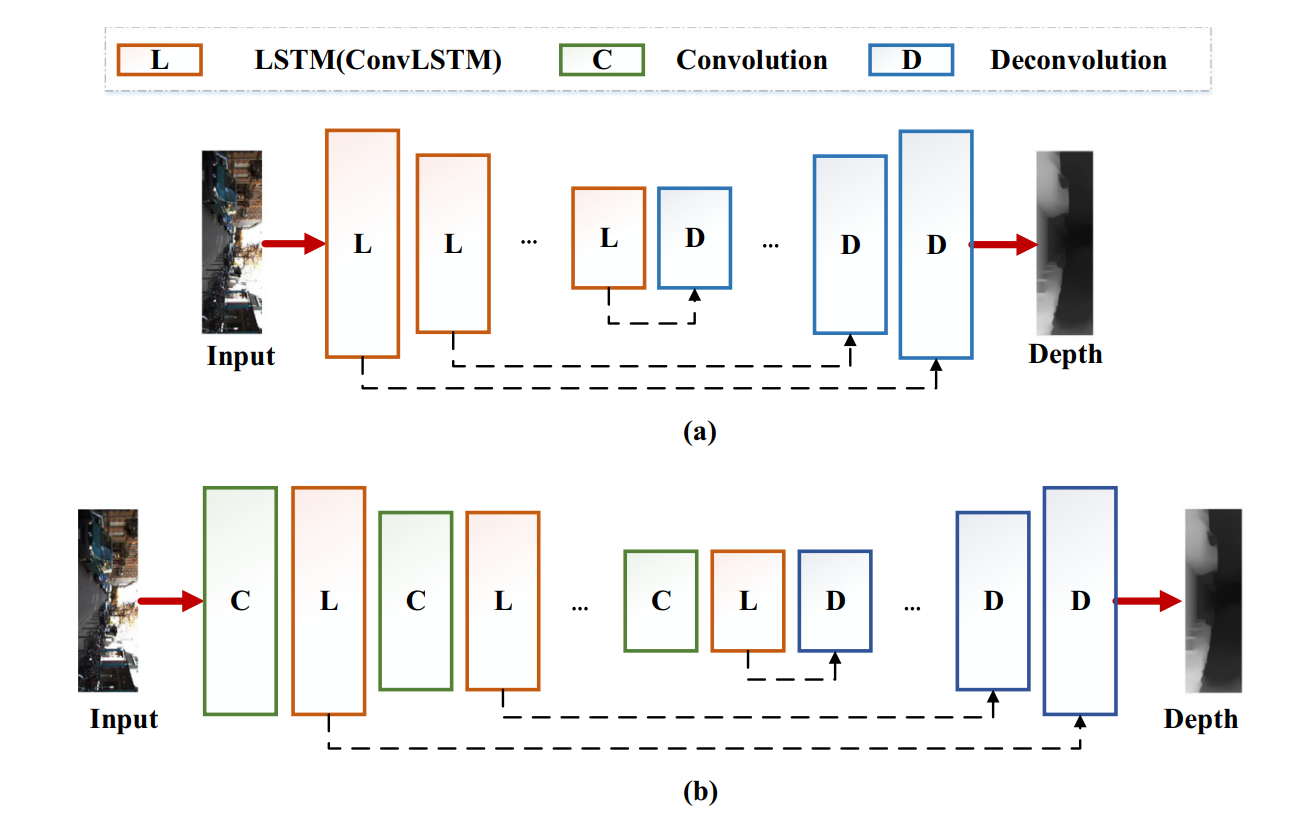}}
\caption{The two major architectures of RNN based networks for predicting monocular depth maps. (a) shows the architecture containing only LSTMs in the encoder while (b) uses convolutions in addition to the LSTMs. Source: \cite{b7}}
\label{fig:image8}
\end{figure}

RNNs are inter-sequence models \cite{b28}\cite{b29} with memory storage capabilities, and are introduced into monocular depth estimation to learn temporal features from video sequences. The RNN contains an input unit, a hidden unit, and three parts: input unit, output unit, and hidden unit. The input of a hidden unit consists of the outputs of both the current input unit and the previously hidden unit.
LSTMs, are a specific type of RNNs where feedback connections allow these networks to learn temporal dependencies between data points. This can be exploited by using video-based datasets where the advancing frames can be fed through the LSTM networks and depth maps are extracted. Fig. \ref{fig:image8} gives an overview of the general pipelines involved in RNN-based depth prediction models where one uses only LSTM (or ConvLSTM), the other uses convolution and LSTM (or ConvLSTM) layers.

RNN and LSTM based networks are employed in the works of \cite{b30} where spatio-temporal consistencies in the datasets are exploited using a Convolutional LSTM, \cite{b31} where attention mechanism of ConvLSTM and ConvGRU are compared, \cite{b32} and \cite{b33}.

\vspace{9pt}

\subsection{Segmentation models for MDE}
Drawing similarities from the pixel-level nature of segmentation in computer vision, monocular depth estimation is a great fit for applying those models for the task of depth estimation. Models for semantic segmentation \cite{b34} specifically work well as they can divide the images into different “stuff” that can have different spatial positions within the field of view and based on that the depth maps can be extracted. The concept of knowledge distillation \cite{b35} about using a large pre-trained “teacher” network to train a smaller “student” network is used in addition to semantic segmentation models like U-Net \cite{b36} and RefineNet \cite{b37} by the works of \cite{b38} and \cite{b39}. The work of \cite{b40} uses a novel segmentation-based learning network to estimate depth in monocular 360-degree videos.

\begin{figure}[htbp]
\centerline{\includegraphics[scale=0.25]{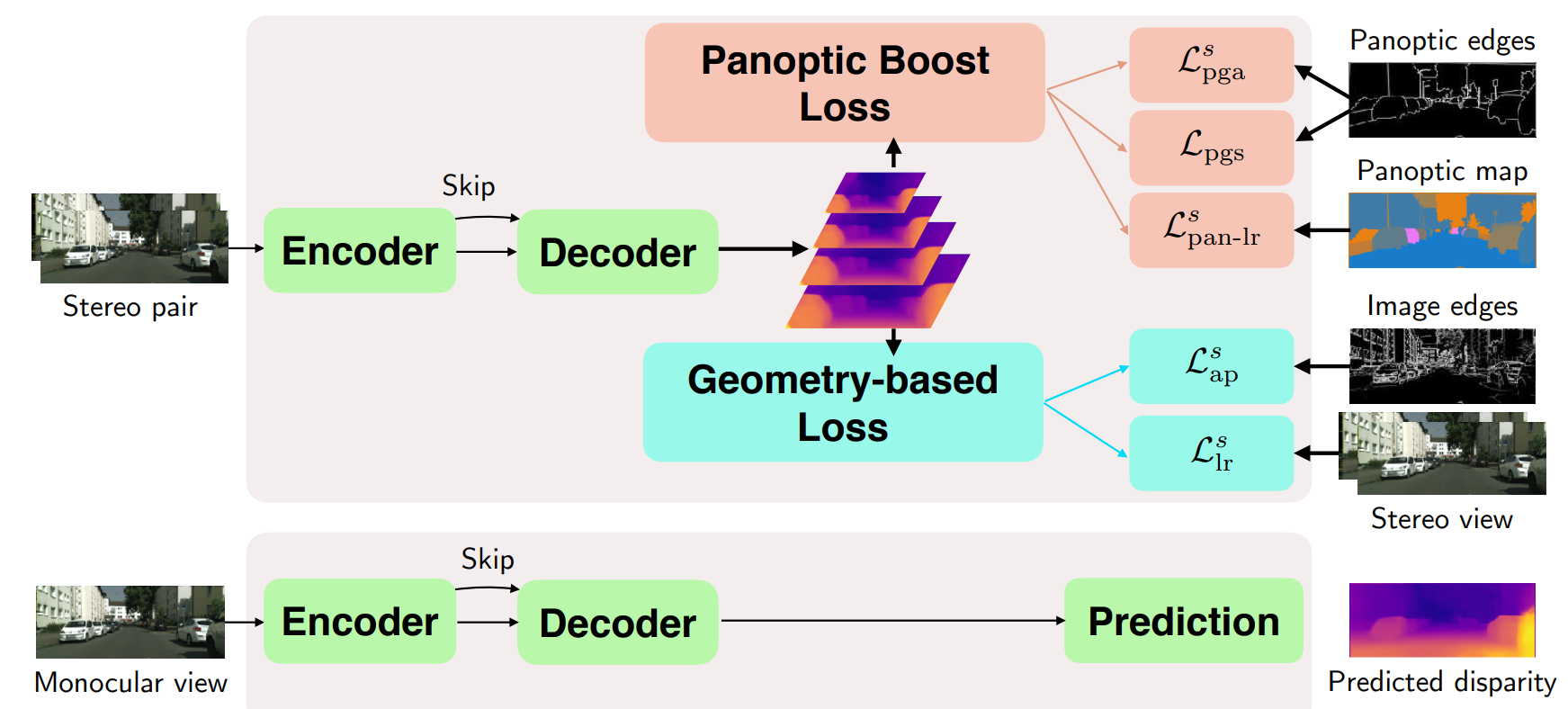}}
\caption{Architectural diagram of our panoptic-boosted monocular depth architecture at training (top) and test time (bottom). During training, our network takes both views as input and produces the corresponding left and right disparity maps in the decoder’s output. At test time, the network uses only a single-view RGB image for prediction. Source: \cite{b42}}
\label{fig:image9}
\end{figure}

Furthermore, the advancements in panoptic segmentation \cite{b41} have a significant potential of making an impact in the field with its segregation of background and instances as “stuff” and “things”, making it intuitive to estimate depth in scenes. \cite{b42} and \cite{b43} make contributions towards MDE using panoptic segmentation. Fig. \ref{fig:image9} shows the model architecture used by \cite{b42} for aiding depth estimation using panoptic segmentation.

\vspace{12pt}

\section{Deep Learning Training Paradigms}

\subsection{Supervised Learning}
\begin{figure}[htbp]
\centerline{\includegraphics[scale=0.30]{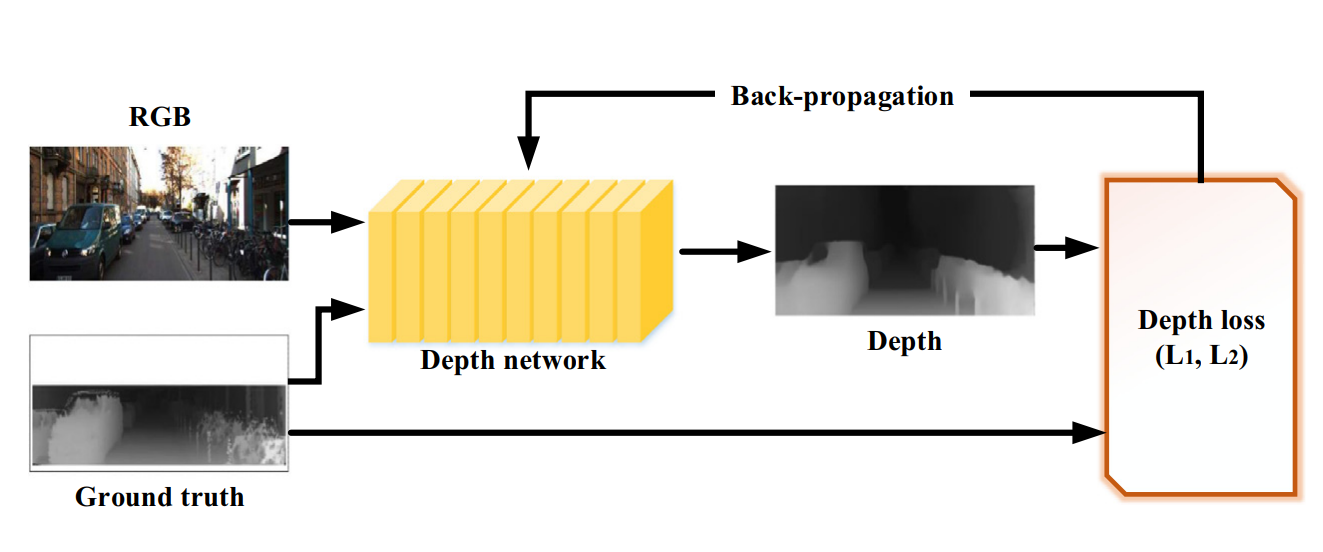}}
\caption{The general architecture of supervised learning for MDE, with the RGB and GT depth images as inputs and depth maps as outputs. Source: \cite{b7}}
\label{fig:image10}
\end{figure}

Supervised learning networks for monocular depth estimation are trained using the Ground Truth depth maps. The purpose of learning is to penalize the error between the prediction and the ground truth depth map on a loss function which is based on the log depth \cite{b44} and the inverse Huber function (Berhu). The goal is for the depth model to converge when the predicted depth value is as close to GT as possible. The generalized pipeline for supervised learning using depth maps as GT is summarized in Fig. \ref{fig:image10}.

Most methods discussed in this paper so-far use a supervised learning approach where the annotated datasets with the GT depth maps are used by a CNN/LSTM/CRF-based learning network for depth estimation, often using stereo pairs for increased accuracy. 

DenseNet \cite{b45} and ResNet \cite{b46} are widely used backbones for CNN-based monocular depth estimation solutions like \cite{b47} and \cite{b48}.

Solutions integrating Conditional Random Fields (CRFs) into supervised learning networks have also yielded state-of-the-art results such as the work of \cite{b49} that divides input images into small windows to make the computation of fully-connected CRFs feasible via the help of a multi-head attention mechanism. 

\vspace{9pt}

\subsection{Unsupervised Learning}
High-resolution, publicly labeled datasets still require a high amount of human resources and intensive work. Therefore, researchers are investigating unsupervised deep learning methods for monocular depth estimation without the use of GT depth maps. Unsupervised monocular depth estimation is typically trained in pairwise stereo images or monocular image sequences and tested in monocular images. 

\begin{figure}[htbp]
\centerline{\includegraphics[scale=0.30]{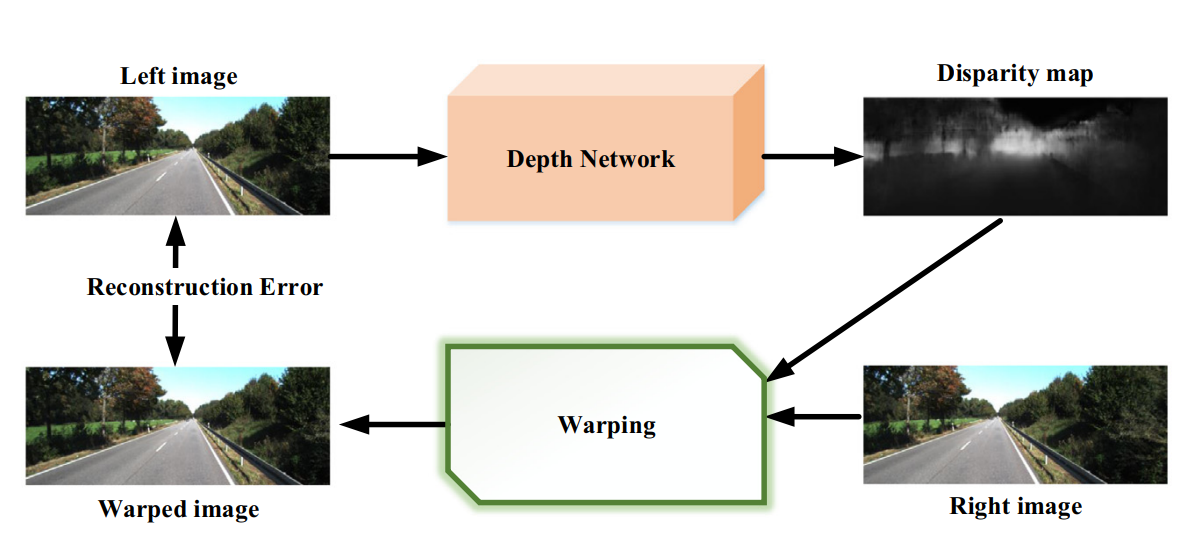}}
\caption{General pipeline for stereo matching based unsupervised learning for depth prediction on monocular sources. Source: \cite{b7}}
\label{fig:image11}
\end{figure}

The two most common ways of training unsupervised MDE models are stereo matching and using monocular sequences. A well-known implementation of the former is \cite{b50} where a traditional belief propagation approach is used to build a depth estimation system. \cite{b51} propose a CNN-based architecture where the model learns depth maps of the right and left views. The work of MonoDepth \cite{b52} with a 2D CNN architecture using an unsupervised learning approach and a combination of disparity smoothness Loss, appearance matching loss, and left-right disparity consistency loss resulted in a significant improvement upon the then SOTA models on the KITTI dataset in 2017. 3D CNNs have also been used by researchers \cite{b5}\cite{b54}\cite{b55}\cite{b56} to take into account the temporal consistency of stereo sequences in addition to the spatial part that is handled by regular CNNs. Fig. \ref{fig:image11} provides the architecture of the general implementation of a stereo matching unsupervised MDE pipeline.

\begin{figure}[htbp]
\centerline{\includegraphics[scale=0.45]{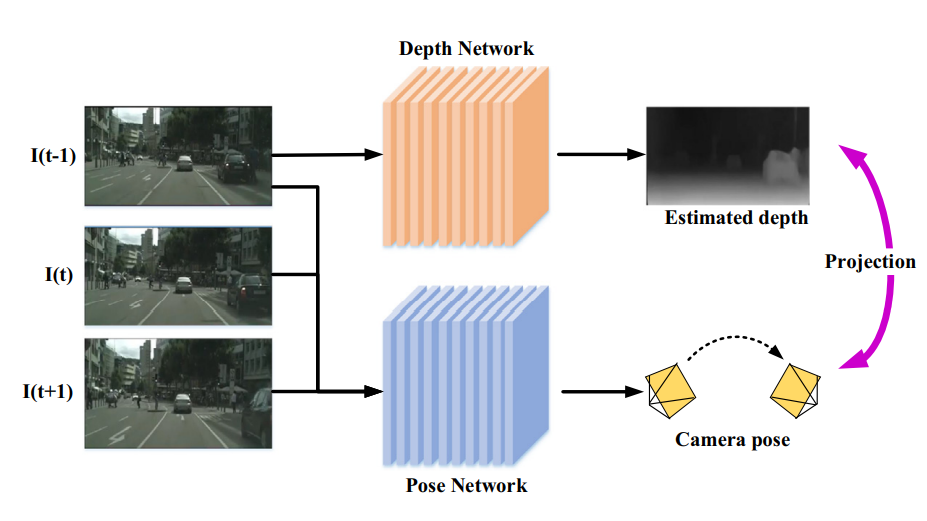}}
\caption{General pipeline for monocular video sequence based unsupervised learning  for depth prediction. Source: \cite{b7}}
\label{fig:image12}
\end{figure}

The other approach to training unsupervised learning-based networks is the use of monocular sequences. This is especially attractive as a research topic due to the higher availability and easier collection process of monocular depth estimation datasets. It also avoids getting into the issues posed by stereo matching relating to projection and left-right source mapping. \cite{b57} and the subsequent related work of \cite{b58} propose methodologies for training unsupervised learning networks on unstructured monocular video sequences along with other elements such as SLAM and optical flow. Fig. \ref{fig:image12} shows the generalized pipeline for this approach.

\vspace{9pt}

\subsection{Semi-supervised Learning}
To effectively use unlabeled datasets available easily in the public domain as well as easy enough to capture by small teams, for improving learning performance, semi-supervised learning approaches are used. These methods can also use other sensor and depth information from sources such as synthetic data, LIDAR, and surface normals, to reduce the model's need for ground truth depth maps, improving the depth map accuracy. The work of \cite{b59} introduces a learning network that works on sparse data along with the RGB data in a stereo-aligned geometric constraint manner. The model then generates two depth maps from these input sources, for which loss is calculated separately with the experiments showing an improvement with this model over a supervised one. \cite{b60} uses a mutual distillation-based loss function in a semi-supervised learning network setting, showing very good performance on KITTI and Cityscapes datasets \cite{b61}.

\vspace{9pt}

\subsection{Self-supervised Learning}
Since the depth value in real-world applications is much larger than the value these neural networks can consistently generate, a proper depth representation will improve the performance considerably. Therefore, the appropriate choice of depth representation to facilitate feature representation learning plays an important role in depth learning and self-supervised monocular motion. SSL (self-supervised learning) is a machine learning technique. It gets its information from unlabeled sample data. It's a kind of learning that's halfway between supervised and unsupervised. 

\begin{figure}[htbp]
\centerline{\includegraphics[scale=0.23]{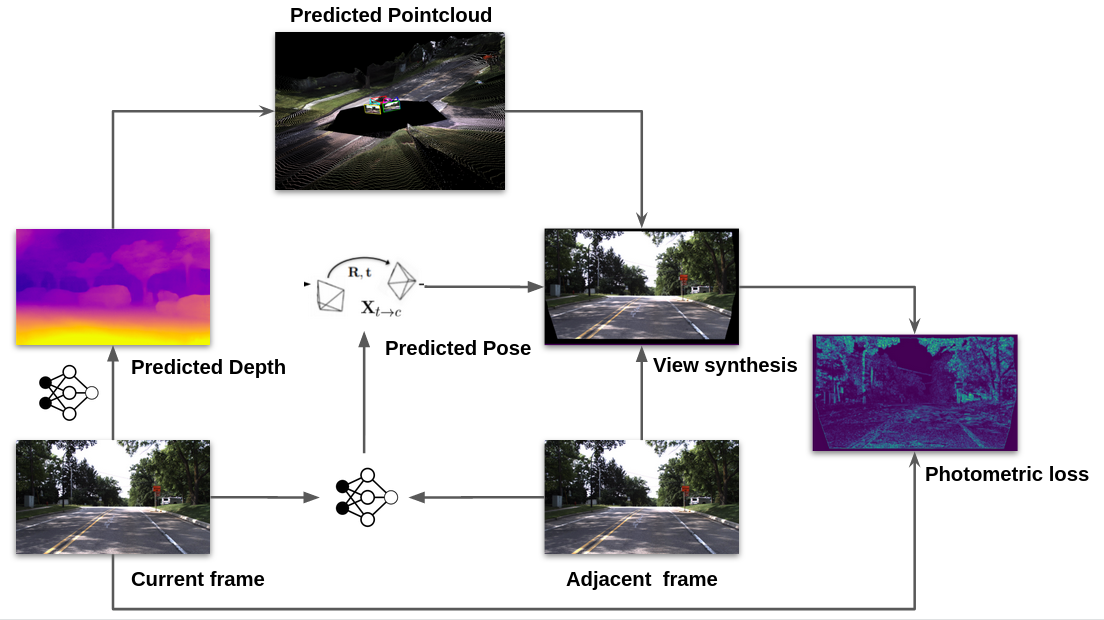}}
\caption{Predicting depth maps using self-supervised learning, a general pipeline}
\label{fig:image13}
\end{figure}

There are two stages to learning. The job is first solved using pseudo-labels, which aid in the initialization of network weights. Second, either supervised or unsupervised learning is used to complete the assignment. In recent years, self-supervised learning has yielded encouraging outcomes. The main benefit of SSL is that it allows training to take place with lower-quality data rather than focusing on improving outcomes. In the case of outdoor depth estimation, the availability of diverse and long-range datasets makes it hard to train generalizable deep learning models. Here self-supervised learning comes into the picture by using the sparse annotated part of the dataset and generating new data to train on. Fig. \ref{fig:image13} visualizes this architecture of SSL based depth estimation.

The popular work of MonoDepth2 \cite{b62} uses a fully convolutional U-Net for depth prediction along with a pose network to account for temporal consistency in video frames. To consider occlusions they utilize per-pixel re-projection with a specific loss function and then upsample the depth maps. SuperDepth \cite{b63} proposes a super-resolution-based depth estimation solution along with a novel augmentation layer that improves prediction accuracy. \cite{b64} introduces a novel technique for SSL-based depth estimation approaches that bring about the then state-of-the-art results by involving uncertainty modeling in this training paradigm. \cite{b65} uses an SSL-based network for depth estimation that is used along with LiDAR data for the task of depth completion in outdoor scenes. Other works such as \cite{b66} are also pushing the envelope of what is possible with the self-supervised learning framework by integrating popular model architectures like transformers that do not require a specific configuration of camera that has to be used for capturing the RGB input.

\vspace{12pt}

\section{Challenges and Trends}
In the last few years, the focus of monocular depth estimation has slightly shifted towards large-scale outdoor and landscape-based data utilization which aligns with the applications they are required for. In this section, we discuss the limitations as well as future research scope that come with them.

\subsection{Collection of long-range datasets}
The applications of autonomous UAVs, robotics, and landscape-level 3D reconstruction require very-long range datasets (captured from 100+ meters far). Most existing research is done for indoor, short-range outdoor, and moving vehicle outdoor datasets. The lack of long-range datasets makes it difficult to extrapolate or adapt the available data too, thus research relating to the collection of such datasets would be an interesting avenue.

Synthetic datasets created using virtual worlds in simulators are the easiest way to approach this problem but care has to be taken in including real-world elements into these datasets such as natural conditions of lighting and haze, occlusion from dynamic objects, and perspective geometry that might be “off” in virtual environments.

\subsection{Integration with semantic segmentation}
There are pieces of work that include segmentation into the overall depth estimation system of supervised and self-supervised learning but these two are treated as independent modules rather than a co-dependent framework. This results in higher computational requirements for training these models and creating/tuning them according to the use case. Research in the direction of integrating these models is a promising area of work.

\subsection{Real-time inference}
The exploitation of temporal consistencies done by 3D CNNs, LSTMs, and other attention mechanisms is still prone to the static nature of per-frame changes in real-world applications that render the task of outdoor depth estimation as a problem of a single image depth estimation even when video sequences are involved. The existing work that balances this issue makes the tradeoff of using deep networks that require a lot of computation resources and time. Research in the area of utilizing lightweight networks similar to what has been done for segmentation is essential to the real-world adoption of these techniques.

\vspace{12pt}

\section{Conclusion}
Outdoor monocular depth estimation is an important step toward the full realization of applications in robotics and simulation. This paper surveys the publicly available datasets, and deep learning methods and summarizes the training approaches used by existing models. Moreover, this paper discusses the performance of popular approaches in different scenarios and the limitations associated with them. In the end, we identify and list the current challenges and related open research opportunities for the task of outdoor depth estimation using monocular vision sources.


\vspace{12pt}

\begin{thebibliography}{00}
\bibitem{b1} G. Eason, B. Noble, and I. N. Sneddon, ``On certain integrals of Lipschitz-Hankel type involving products of Bessel functions,'' Phil. Trans. Roy. Soc. London, vol. A247, pp. 529--551, April 1955.

\bibitem{b1} Liu, Yang, et al. "A survey of depth estimation based on computer vision." 2020 IEEE Fifth International Conference on Data Science in Cyberspace (DSC). IEEE, 2020.
 
\bibitem{b2} Chen, Hui, et al. "3D reconstruction approach for outdoor scene based on multiple point cloud fusion." Journal of the Indian Society of Remote Sensing 47.10 (2019): 1761-1772.

\bibitem{b3} Lan, Ziquan, Zi Jian Yew, and Gim Hee Lee. "Robust point cloud based reconstruction of large-scale outdoor scenes." Proceedings of the IEEE/CVF Conference on Computer Vision and Pattern Recognition. 2019.

\bibitem{b4} Tsai, Yi-Min, Yu-Lin Chang, and Liang-Gee Chen. "Block-based vanishing line and vanishing point detection for 3D scene reconstruction." 2006 international symposium on intelligent signal processing and communications. IEEE, 2006.

\bibitem{b5} Tang, Chang, Chunping Hou, and Zhanjie Song. "Depth recovery and refinement from a single image using defocus cues." Journal of Modern Optics 62.6 (2015): 441-448.

\bibitem{b6} Zhang, Ping, et al. "Stereoscopic video saliency detection based on spatiotemporal correlation and depth confidence optimization." Neurocomputing 377 (2020): 256-268.

\bibitem{b7} Ming, Yue, et al. "Deep learning for monocular depth estimation: A review." Neurocomputing 438 (2021): 14-33.

\bibitem{b8} Lopes, Alexandre, Roberto Souza, and Helio Pedrini. "A Survey on RGB-D Datasets." arXiv preprint arXiv:2201.05761 (2022).

\bibitem{b9} Geiger, Andreas, et al. "Vision meets robotics: The kitti dataset." The International Journal of Robotics Research 32.11 (2013): 1231-1237.

\bibitem{b10} Saxena, Ashutosh, Min Sun, and Andrew Y. Ng. "Make3d: Learning 3d scene structure from a single still image." IEEE transactions on pattern analysis and machine intelligence 31.5 (2008): 824-840.

\bibitem{b11} Ramezani, Milad, et al. "The newer college dataset: Handheld lidar, inertial and vision with ground truth." 2020 IEEE/RSJ International Conference on Intelligent Robots and Systems (IROS). IEEE, 2020.

\bibitem{b12} Li, Zhengqi, and Noah Snavely. "Megadepth: Learning single-view depth prediction from internet photos." Proceedings of the IEEE Conference on Computer Vision and Pattern Recognition. 2018.

\bibitem{b13} Vasiljevic, Igor, et al. "Diode: A dense indoor and outdoor depth dataset." arXiv preprint arXiv:1908.00463 (2019).

\bibitem{b14} Yang, Guorun, et al. "Drivingstereo: A large-scale dataset for stereo matching in autonomous driving scenarios." Proceedings of the IEEE/CVF Conference on Computer Vision and Pattern Recognition. 2019.

\bibitem{b15} Tateno, Keisuke, Nassir Navab, and Federico Tombari. "Distortion-aware convolutional filters for dense prediction in panoramic images." Proceedings of the European Conference on Computer Vision (ECCV). 2018.

\bibitem{b16} Yuan, Weihao, et al. "NeW CRFs: Neural Window Fully-connected CRFs for Monocular Depth Estimation." arXiv preprint arXiv:2203.01502 (2022).

\bibitem{b17} Zhou, Keyang, Kaiwei Wang, and Kailun Yang. "PADENet: An efficient and robust panoramic monocular depth estimation network for outdoor scenes." 2020 IEEE 23rd International Conference on Intelligent Transportation Systems (ITSC). IEEE, 2020.

\bibitem{b18} Zhuang, Chuanqing, et al. "ACDNet: Adaptively Combined Dilated Convolution for Monocular Panorama Depth Estimation." arXiv preprint arXiv:2112.14440 (2021).

\bibitem{b19} Chao, Fang-Yi, et al. "A Multi-FoV Viewport-Based Visual Saliency Model Using Adaptive Weighting Losses for 360$^\circ $ Images." IEEE Transactions on Multimedia 23 (2020): 1811-1826.

\bibitem{b20} Schops, Thomas, et al. "A multi-view stereo benchmark with high-resolution images and multi-camera videos." Proceedings of the IEEE Conference on Computer Vision and Pattern Recognition. 2017.

\bibitem{b21} Mancini, Michele, et al. "Toward domain independence for learning-based monocular depth estimation." IEEE Robotics and Automation Letters 2.3 (2017): 1778-1785.

\bibitem{b22} Reza, Md Alimoor, Jana Kosecka, and Philip David. "FarSight: Long-Range Depth Estimation from Outdoor Images." 2018 IEEE/RSJ International Conference on Intelligent Robots and Systems (IROS). IEEE, 2018.

\bibitem{b23} de La Garanderie, Greire Payen, Amir Atapour Abarghouei, and Toby P. Breckon. "Eliminating the blind spot: Adapting 3d object detection and monocular depth estimation to 360 panoramic imagery." Proceedings of the European Conference on Computer Vision (ECCV). 2018.

\bibitem{b24} Zhu, Jun-Yan, et al. "Unpaired image-to-image translation using cycle-consistent adversarial networks." Proceedings of the IEEE international conference on computer vision. 2017.

\bibitem{b25} Liu, Fayao, Chunhua Shen, and Guosheng Lin. "Deep convolutional neural fields for depth estimation from a single image." Proceedings of the IEEE conference on computer vision and pattern recognition. 2015.

\bibitem{b26} Bhat, Shariq Farooq, Ibraheem Alhashim, and Peter Wonka. "Adabins: Depth estimation using adaptive bins." Proceedings of the IEEE/CVF Conference on Computer Vision and Pattern Recognition. 2021.

\bibitem{b27} Garg, Ravi, et al. "Unsupervised cnn for single view depth estimation: Geometry to the rescue." European conference on computer vision. Springer, Cham, 2016.

\bibitem{b28} Ceni, Andrea, Peter Ashwin, and Lorenzo Livi. "Interpreting recurrent neural networks behaviour via excitable network attractors." Cognitive Computation 12.2 (2020): 330-356.

\bibitem{b29} Gregor, Karol, et al. "Draw: A recurrent neural network for image generation." International Conference on Machine Learning. PMLR, 2015.

\bibitem{b30} Zhang, Haokui, et al. "Exploiting temporal consistency for real-time video depth estimation." Proceedings of the IEEE/CVF International Conference on Computer Vision. 2019.

\bibitem{b31} Maslov, Dmitrii, and Ilya Makarov. "Online supervised attention-based recurrent depth estimation from monocular video." PeerJ Computer Science 6 (2020): e317.

\bibitem{b32} CS Kumar, Arun, Suchendra M. Bhandarkar, and Mukta Prasad. "Depthnet: A recurrent neural network architecture for monocular depth prediction." Proceedings of the IEEE Conference on Computer Vision and Pattern Recognition Workshops. 2018.

\bibitem{b33} Wang, Rui, Stephen M. Pizer, and Jan-Michael Frahm. "Recurrent neural network for (un-) supervised learning of monocular video visual odometry and depth." Proceedings of the IEEE/CVF Conference on Computer Vision and Pattern Recognition. 2019.

\bibitem{b34} Long, Jonathan, Evan Shelhamer, and Trevor Darrell. "Fully convolutional networks for semantic segmentation." Proceedings of the IEEE conference on computer vision and pattern recognition. 2015.

\bibitem{b35} Hinton, Geoffrey, Oriol Vinyals, and Jeff Dean. "Distilling the knowledge in a neural network." arXiv preprint arXiv:1503.02531 2.7 (2015).

\bibitem{b36} Ronneberger, Olaf, Philipp Fischer, and Thomas Brox. "U-net: Convolutional networks for biomedical image segmentation." International Conference on Medical image computing and computer-assisted intervention. Springer, Cham, 2015.

\bibitem{b37} Lin, Guosheng, et al. "Refinenet: Multi-path refinement networks for high-resolution semantic segmentation." Proceedings of the IEEE conference on computer vision and pattern recognition. 2017.

\bibitem{b38} Cho, Jaehoon, et al. "Deep monocular depth estimation leveraging a large-scale outdoor stereo dataset." Expert Systems with Applications 178 (2021): 114877.

\bibitem{b39} Nekrasov, Vladimir, et al. "Real-time joint semantic segmentation and depth estimation using asymmetric annotations." 2019 International Conference on Robotics and Automation (ICRA). IEEE, 2019.

\bibitem{b40} Feng, Qi, Hubert PH Shum, and Shigeo Morishima. "360 Depth Estimation in the Wild-the Depth360 Dataset and the SegFuse Network." 2022 IEEE Conference on Virtual Reality and 3D User Interfaces (VR). IEEE, 2022.

\bibitem{b41} Yuan, Haobo, et al. "PolyphonicFormer: Unified Query Learning for Depth-aware Video Panoptic Segmentation." arXiv preprint arXiv:2112.02582 (2021).

\bibitem{b42} Saeedan, Faraz, and Stefan Roth. "Boosting monocular depth with panoptic segmentation maps." Proceedings of the IEEE/CVF Winter Conference on Applications of Computer Vision. 2021.

\bibitem{b43} Qiao, Siyuan, et al. "Vip-deeplab: Learning visual perception with depth-aware video panoptic segmentation." Proceedings of the IEEE/CVF Conference on Computer Vision and Pattern Recognition. 2021.

\bibitem{b44} Eigen, David, and Rob Fergus. "Predicting depth, surface normals and semantic labels with a common multi-scale convolutional architecture." Proceedings of the IEEE international conference on computer vision. 2015.

\bibitem{b45} Huang, Gao, et al. "Densely connected convolutional networks." Proceedings of the IEEE conference on computer vision and pattern recognition. 2017.

\bibitem{b46} He, Kaiming, et al. "Deep residual learning for image recognition." Proceedings of the IEEE conference on computer vision and pattern recognition. 2016.

\bibitem{b47} Laina, Iro, et al. "Deeper depth prediction with fully convolutional residual networks." 2016 Fourth international conference on 3D vision (3DV). IEEE, 2016.

\bibitem{b48} Lee, Jin Han, et al. "From big to small: Multi-scale local planar guidance for monocular depth estimation." arXiv preprint arXiv:1907.10326 (2019).

\bibitem{b49} Zhang, Zhenyu, et al. "Progressive hard-mining network for monocular depth estimation." IEEE Transactions on Image Processing 27.8 (2018): 3691-3702.

\bibitem{b50} Sun, Jian, Nan-Ning Zheng, and Heung-Yeung Shum. "Stereo matching using belief propagation." IEEE Transactions on pattern analysis and machine intelligence 25.7 (2003): 787-800.

\bibitem{b51} Garg, Ravi, et al. "Unsupervised cnn for single view depth estimation: Geometry to the rescue." European conference on computer vision. Springer, Cham, 2016.

\bibitem{b52} Godard, Clément, Oisin Mac Aodha, and Gabriel J. Brostow. "Unsupervised monocular depth estimation with left-right consistency." Proceedings of the IEEE conference on computer vision and pattern recognition. 2017.

\bibitem{b53} Chen, Chuangrong, Xiaozhi Chen, and Hui Cheng. "On the over-smoothing problem of cnn based disparity estimation." Proceedings of the IEEE/CVF International Conference on Computer Vision. 2019.

\bibitem{b54} Tulyakov, Stepan, Anton Ivanov, and Francois Fleuret. "Practical deep stereo (pds): Toward applications-friendly deep stereo matching." Advances in neural information processing systems 31 (2018).

\bibitem{b55} Kendall, Alex, et al. "End-to-end learning of geometry and context for deep stereo regression." Proceedings of the IEEE international conference on computer vision. 2017.

\bibitem{b56} Chang, Jia-Ren, and Yong-Sheng Chen. "Pyramid stereo matching network." Proceedings of the IEEE conference on computer vision and pattern recognition. 2018.

\bibitem{b57} A. Moreau, M. Mancas and T. Dutoit, "Unsupervised depth prediction from monocular sequences: Improving performances through instance segmentation," 2020 17th Conference on Computer and Robot Vision (CRV), 2020

\bibitem{b58} Wang, Guangming, et al. "Unsupervised learning of depth, optical flow and pose with occlusion from 3d geometry." IEEE Transactions on Intelligent Transportation Systems 23.1 (2020): 308-320.

\bibitem{b59} Kuznietsov, Yevhen, Jorg Stuckler, and Bastian Leibe. "Semi-supervised deep learning for monocular depth map prediction." Proceedings of the IEEE conference on computer vision and pattern recognition. 2017.

\bibitem{b60} Baek, Jongbeom, Gyeongnyeon Kim, and Seungryong Kim. "Semi-Supervised Learning with Mutual Distillation for Monocular Depth Estimation." arXiv preprint arXiv:2203.09737 (2022).

\bibitem{b61} Cordts, Marius, et al. "The cityscapes dataset for semantic urban scene understanding." Proceedings of the IEEE conference on computer vision and pattern recognition. 2016.

\bibitem{b62} Godard, Clément, et al. "Digging into self-supervised monocular depth estimation." Proceedings of the IEEE/CVF International Conference on Computer Vision. 2019.

\bibitem{b63} Pillai, Sudeep, Rareş Ambruş, and Adrien Gaidon. "Superdepth: Self-supervised, super-resolved monocular depth estimation." 2019 International Conference on Robotics and Automation (ICRA). IEEE, 2019.

\bibitem{b64} Poggi, Matteo, et al. "On the uncertainty of self-supervised monocular depth estimation." Proceedings of the IEEE/CVF Conference on Computer Vision and Pattern Recognition. 2020.

\bibitem{b65} Ma, Fangchang, Guilherme Venturelli Cavalheiro, and Sertac Karaman. "Self-supervised sparse-to-dense: Self-supervised depth completion from lidar and monocular camera." 2019 International Conference on Robotics and Automation (ICRA). IEEE, 2019.

\bibitem{b66} Varma, Arnav, et al. "Transformers in Self-Supervised Monocular Depth Estimation with Unknown Camera Intrinsics." arXiv preprint arXiv:2202.03131 (2022).

\end{thebibliography}
\end{document}